\documentclass[epsf, twocolumn]{IEEEtran}
\usepackage{graphicx}

\begin{document}

\title{Flexible Camera Calibration Using a New Analytical Radial Undistortion Formula with Application to Mobile Robot Localization}

\author{Lili Ma, {\it Student Member, IEEE}, YangQuan Chen and Kevin L. Moore, {\it Senior Members, IEEE}\\ Center for Self-Organizing and Intelligent Systems (CSOIS),\\Dept. of Electrical and Computer Engineering, 4160 Old Main Hill,\\ Utah State University (USU), Logan, UT 84322-4160, USA.\\
Emails: \texttt{lilima@cc.usu.edu, \{yqchen, moorek\}@ece.usu.edu}
}

\maketitle{}

\begin{abstract}
Most algorithms in 3D computer vision rely on the pinhole camera model because of its simplicity, whereas virtually all imaging devices introduce certain amount of nonlinear distortion, where the radial distortion is the most severe part. Common approach to radial distortion is by the means of polynomial approximation, which introduces distortion-specific parameters into the camera model and requires estimation of these distortion parameters. The task of estimating radial distortion is to find a radial distortion model that allows easy undistortion as well as satisfactory accuracy. This paper presents a new radial distortion model with an  easy analytical undistortion formula, which also belongs to the polynomial approximation category. Experimental results are presented to show that with this radial distortion model, satisfactory accuracy is achieved. An application of the new radial distortion model is non-iterative yellow line alignment with a calibrated camera on ODIS, a robot built in our CSOIS (See Fig. \ref{fig: odis1}).
\end{abstract}

\thispagestyle{empty}
\section{Introduction}


\subsection{Related Work: Camera Calibration}
Depending on what kind of calibration object used, there are mainly two categories of calibration methods: photogrammetric calibration and self-calibration. Photogrammetric calibration refers to those methods that observe a calibration object whose geometry in 3-D space is known with a very good precision \cite{Emanuele98introductorycomputervision}. Self-calibration does not need any calibration object. It only requires point matches from image sequence. In \cite{Faugeras92self-calibration}, it is shown that it is possible to calibrate a camera just by pointing it to the environment, selecting points of interest and then tracking them in the image as the camera moves. The obvious advantage of the self-calibration method is that it is not necessary to know the camera motion and it is easy to set up. The disadvantage is that it is usually considered unreliable \cite{Bougnoux98criticismself-calibration}. A four step calibration procedure is proposed in \cite{Heikkil97fourstepcameracalibration} where the calibration is performed with a known 3D target. The four steps in \cite{Heikkil97fourstepcameracalibration} are: linear parameter estimation, nonlinear optimization, correction using circle/ellipse, and image correction. But for a simple start, linear parameter estimation and nonlinear optimization are enough. In \cite{STURM99planebasedcalibrationsigularities}, a plane-based calibration method is described where the calibration is performed by first determining the absolute conic ${\bf B} = {\bf A}^{-T} {\bf A}^{-1}$, where $\bf A$ is a matrix formed by the camera's intrinsic parameters. In \cite{STURM99planebasedcalibrationsigularities}, the parameter $\gamma$ (a parameter describing the skewness of the two image axes) is assumed to be zero and it is observed that only the relative orientations of planes and camera are of importance in avoiding singularities because the planes that are parallel to each other provide exactly the same information. The camera calibration method in \cite{zhang99calibrationinpaper,zhang99calibrationinreport} is regarded as a great contribution to the camera calibration. It focuses on the desktop vision system and advances 3D computer vision one step from laboratory environments to the real world. The proposed method in \cite{zhang99calibrationinpaper,zhang99calibrationinreport} lies between the photogrammetric calibration and the self-calibration, because 2D metric information is used rather than 3D. The key feature of the calibration method in \cite{zhang99calibrationinpaper,zhang99calibrationinreport} is that the absolute conic $\bf B$ is used to estimate the intrinsic parameters and the parameter $\gamma$ can be considered. The proposed technique in \cite{zhang99calibrationinpaper,zhang99calibrationinreport} only requires the camera to observe a planar pattern at a few (at least 3, if both the intrinsic and the extrinsic parameters are to be estimated uniquely) different orientations. Either the camera or the calibration object can be moved by hand as long as they cause no singularity problem and the motion of the calibration object or camera itself needs not to be known in advance.

After estimation of camera parameters, a projection matrix $\bf M$ can directly link a point in the 3-D world reference frame to its projection (undistorted) in the image plane. That is
\begin{eqnarray}
\label{eqn: projection matrix}
\lambda \left [\matrix{u \cr v \cr 1} \right ] &=& {\bf M} \left [\matrix{X^w \cr Y^w \cr Z^w \cr 1} \right ] = {\bf A} \left [\matrix{{\bf R} & {\bf t}}\right] \left [\matrix{X^w \cr Y^w \cr Z^w \cr 1} \right ] \nonumber\\
&=& \left [ \matrix { \alpha & \gamma &u_0 \cr 0 & \beta & v_0 \cr 0 & 0 & 1 } \right ] \left [\matrix{{\bf R} & {\bf t}}\right] \left [\matrix{ X^w \cr Y^w \cr Z^w \cr 1} \right ],
\end{eqnarray}
where $\lambda$ is an arbitrary scaling factor and the matrix $\bf A$ fully depends on the 5 intrinsic parameters with their detail descriptions in Table \ref{table: variables used}, where some other variables used throughout this paper are also listed.

The calibration method used in this work is to first estimate the projection matrix and then use the absolute conic to estimate the intrinsic parameters \cite{zhang99calibrationinpaper,zhang99calibrationinreport}. The detail procedures are summarized below:
\begin{itemize}
\item{Linear Parameter Estimation,}
    \begin{itemize}
    \item Estimation of Intrinsic Parameters;
    \item Estimation of Extrinsic Parameters;
    \item Estimation of Distortion Coefficients;
    \end{itemize}
\item Nonlinear Optimization.
\end{itemize}

\begin{table}[htb]
\centering
\caption{List of Variables}
\label{table: variables used}
{\small
{\begin {tabular}{|c|l|}\hline
{\bf Variable} & {\bf Description} \\[1ex]\hline
$P^w = [X^w, Y^w, Z^w]^T$       & 3-D point in world frame\\[1ex]\hline
$P^c = [X^c, Y^c, Z^c]^T$       & 3-D point in camera frame \\[1ex]\hline
${\bf k} = (k_1, \, k_2)$        & Distortion coefficients \\[1ex]\hline
$(u_d, \, v_d)$              & Distorted image points \\[1ex]\hline
$(u, \, v)$                  & Undistorted image points \\[1ex]\hline
$(x_d, \, y_d)$              & $\left[ \matrix{x_d\cr y_d \cr 1}\right] = A^{-1} \left[ \matrix{u_d \cr v_d \cr 1}\right]$\\[3.2ex]
\hline
$(x, \, y)$                  & $\left[ \matrix{x \cr y \cr 1}\right] = A^{-1} \left[ \matrix{u \cr v \cr 1}\right]$ \\[4ex]\hline
$r$                     & $r^2 = x^2 + y^2$ \\[1ex]\hline
$\alpha, \beta, \gamma, u_0, v_0$   & 5 intrinsic parameters \\[1ex]\hline
$J$                         & Objective function \\[1ex]\hline
${\bf A} = \left [ \matrix {
\alpha & \gamma &u_0 \cr
0 & \beta & v_0 \cr
0 & 0 & 1
} \right ]$                 & Camera intrinsic matrix \\[4ex]\hline
$\bf B = {\bf A}^{-T} {\bf A}^{-1}$ & Absolute conic \\[1ex]\hline
$\bf M$                     & Projection matrix \\[1ex]\hline
\end {tabular}}
}
\end{table}

\subsection{Radial Distortion}
Radial distortion causes an inward or outward displacement of a given image point from its ideal location. The negative radial displacement of the image points is referred to as the barrel distortion, while the positive radial displacement is referred to as the pincushion distortion \cite{Juyang92distortionmodel}. The radial distortion is governed by the following equation \cite{zhang99calibrationinpaper,Juyang92distortionmodel}:
\begin{equation}
\label{eqn: general distortion model}
F(r) = r \, f(r) = r \, (1 + k_1 r^2 + k_2 r^4 + k_3 r^6+ \cdots),
\end{equation}
where $k_1, k_2, k_3, \ldots$ are the distortion coefficients and $r^2 = x^2 + y^2$ with $(x,y)$ the normalized undistorted projected points in the camera frame. The distortion is usually dominated by the radial components, and especially dominated by the first term. It has also been found that too high an order in (\ref{eqn: general distortion model}) may cause numerical instability \cite{zhang99calibrationinreport,tsai87AVersatile,GWei94Implicit}. In this paper, at most two terms of radial distortion are considered. When using two coefficients, the relationship between the distorted and the undistorted image points becomes \cite{zhang99calibrationinpaper}
\begin{eqnarray}
\label{eqn: radial distortion order 2 4}
u_d - u_0 &=& (u-u_0) \, (1 + k_1 r^2 + k_2 r^4) \nonumber\\
v_d - v_0 &=& (v-v_0) \, (1 + k_1 r^2 + k_2 r^4).
\end{eqnarray}
When using two distortion coefficients to model radial distortion as in \cite{zhang99calibrationinpaper,chapter4}, the inverse of the polynomial function in (\ref{eqn: radial distortion order 2 4}) is difficult to perform analytically. In \cite{chapter4}, the inverse function is obtained numerically via an iterative scheme. In \cite{Undistortionchapter}, for practical purpose, only one distortion coefficient $k_1$ is used. Besides the polynomial approximation method mentioned above, a technique for blindly removing lens distortion in the absence of any calibration information in the frequency domain is presented in \cite{Hany01blindremoval}. However, the accuracy reported in \cite{Hany01blindremoval} is by no means comparable to that based on calibration and this approach can be useful in areas where only qualitative results are required. The new radial distortion model proposed in this paper belongs to the polynomial approximation category.

The rest of the paper is organized as follows. Sec.~\ref{sec: Radial Distortion Models} describes the new radial distortion model and its inverse undistortion analytical formula. Experimental results and comparison with existing models are presented in Sec.~\ref{sec: Comparison Results}. One direct application of this new distortion model is discussed in Sec.~\ref{sec: Application}. Finally, some concluding remarks are given in Sec.~\ref{sec: Conclusion}.

\thispagestyle{empty}
\section{Radial Distortion Models}
\label{sec: Radial Distortion Models}
In this paper, we focus on the distortion models while the intrinsic parameters and the extrinsic parameters are achieved using the method presented in \cite{zhang99calibrationinpaper,zhang99calibrationinreport}. According to the radial distortion model in (\ref{eqn: radial distortion order 2 4}), the radial distortion can be resulted in one of the  following two ways:
\begin{itemize}

\item{Transform from the camera frame to the image plane, then perform distortion in the image plane}
\begin{eqnarray}
{\left [\matrix {
x \cr
y}\right]} \rightarrow
{\left [\matrix {
u \cr
v}\right]} \rightarrow
{\left [\matrix {
u_d \cr
v_d}\right];}\nonumber
\end{eqnarray}

\item{Perform distortion in the camera frame, then transform to the image plane}
\begin{eqnarray}
{\left [\matrix {
x \cr
y}\right]} \rightarrow
{\left [\matrix {
x_d \cr
y_d}\right]} \rightarrow
{\left [\matrix {
u_d \cr
v_d}\right],} \nonumber
\end{eqnarray}
where
\begin{eqnarray}
\label{eqn: xdyd and xy}
x_d = x \, f(r), \quad y_d = y \, f(r).
\end {eqnarray}
\end{itemize}
Since
\begin{eqnarray}
\left[ \matrix{u \cr v \cr 1}\right] = {\bf A} \left[ \matrix{x \cr y \cr 1}\right] = \left [ \matrix {
\alpha & \gamma &u_0 \cr
0 & \beta & v_0 \cr
0 & 0 & 1
} \right ] \left[ \matrix{x \cr y \cr 1}\right], \nonumber
\end{eqnarray}
(\ref{eqn: radial distortion order 2 4}) becomes
\begin{eqnarray}
u_d &=& (u-u_0) \, f(r) + u_0 \nonumber \\
    &=& \alpha \, x f(r) + \gamma \, y f(r) + u_0 \nonumber\\
    &=& \alpha \, x_d + \gamma \, y_d + u_0, \nonumber\\
v_d &=& (v-v_0) \, f(r) + v_0\nonumber\\
    &=& \beta \, y_d  + v_0.
\end{eqnarray}
Therefore, it is also true that
\begin{eqnarray}
\left[ \matrix{u_d \cr v_d \cr 1}\right] = {\bf A} \left[ \matrix{x_d \cr y_d \cr 1}\right]. \nonumber
\end{eqnarray}
Thus, the distortion performed in the image plane can also be understood as introducing distortion in the camera frame and then transform back to the image plane.

\subsection{The Existing Radial Distortion Models}
Radial undistortion is to extract $(u,v)$ from $(u_d, v_d)$, which can also be accomplished by extracting $(x,y)$ from $(x_d, y_d)$. The following derivation shows the problem when trying to extract $(x,y)$ from $(x_d, y_d)$ using two distortion coefficients $k_1$ and $k_2$ in (\ref{eqn: radial distortion order 2 4}).

From $(u_d, v_d)$, we can calculate $(x_d, y_d)$ by
\begin{eqnarray}
{\left [\matrix {
x_d \cr
y_d \cr
1}\right]} = {\bf A}^{-1} {\left [\matrix {
u_d \cr
v_d \cr
1}\right]}
= {\left [\matrix {
\frac{1}{\alpha} & -\frac{\gamma}{\alpha \beta} & -\frac{u_0}{\alpha}+\frac{v_0 \gamma}{\alpha \beta} \cr
0 & \frac{1}{\beta} & -\frac{v_0}{\beta} \cr
0 & 0 & 1 }\right]} {\left [\matrix {
u_d \cr
v_d \cr
1}\right],}
\end{eqnarray}
where the camera intrinsic matrix $\bf A$ is invertible by nature. Now, the problem becomes to extracting $(x, y)$ from $(x_d, y_d)$. According to (\ref{eqn: xdyd and xy}),
\begin{eqnarray}
x_d = x f(r) = x [1+k_1(x^2+y^2) + k_2(x^2+y^2)^2]  \nonumber\\
y_d = y f(r) = y [1+k_1(x^2+y^2) + k_2(x^2+y^2)^2].
\end{eqnarray}
It is obvious that $x_d = 0$ iff $x = 0$. When $x_d \neq 0$, by letting $c = y_d/x_d = y /x$, we have $y = cx$ where $c$ is a constant. Substituting $y = cx$ into the above equation gives
\begin{eqnarray}
\label{eqn: undistortion case(1)}
x_d &=& x \, [1+k_1(x^2+c^2x^2)+k_2(x^2+c^2x^2)^2] \nonumber\\
    &=& x + k_1 (1+c^2)x^3+k_2(1+c^2)^2x^5 .
\end{eqnarray}
Let $f(x) = x + k_1 (1+c^2)x^3+k_2(1+c^2)^2x^5$. Then $f(-x) = - f(x)$ and $f(x)$ is an odd function. The analytical solution of (\ref{eqn: undistortion case(1)}) is not a trivial task. This analytical problem is still open (of course, we can use numerical method to solve it). But if we set $k_2 = 0$, the analytical solution is available and the radial undistortion can be done easily. In \cite{Undistortionchapter}, for the same practical reason, only one distortion coefficient $k_1$ is used to approximate the radial distortion, in which case we would expect to see performance degradation. In Sec.~\ref{sec: Comparison Results}, experimental results are presented to show the performance comparison for the cases when $k_2 = 0$ and $k_2 \neq 0$ using the calibrated parameters of three different cameras. Recall that the initial guess for radial distortion is done after having estimated all other parameters (including both intrinsic and extrinsic parameters) and just before the nonlinear optimization step. So, we can reuse the estimated parameters and choose the initial guess for $k_2$ to be 0 and compare the values of objective function after nonlinear optimization.

The objective function used for nonlinear optimization is \cite{zhang99calibrationinpaper}:
\begin{equation}
\label{eqn: objective function}
J = \sum_{i=1}^N \sum_{j=1}^n ||m_{ij}-\hat m({\bf A},k_1, k_2, {\bf R}_i, {\bf t}_i, M_j)||^2,
\end{equation}
where $\hat m({\bf A},k_1, k_2, {\bf R}_i, {\bf t}_i, M_j)$ is the projection of point $M_j$ in the $i^{th}$ image using the estimated parameters and $M_j$ is the $j^{th}$  3D point in the world frame with $Z^w = 0$. Here, $n$ is the number of feature points in the coplanar calibration object and $N$ is the number of images taken for calibration.

\subsection{The New Radial Distortion Model}
\label{sec: new distortion model}
Our new radial distortion model is proposed as:
\begin{equation}
\label{eqn: new distortion model}
F(r) = r \, f(r) = r \, (1 + k_1 r + k_2 r^2),
\end{equation}
which is also a function only related to radius $r$.  The motivation of choosing this radial distortion model is that the resultant approximation of $x_d$ is also an odd function of $x$, as can be seen next.
For $F(r) = r f(r) = r (1 + k_1 r + k_2 r^2)$, we have
\begin{eqnarray}
\begin{array}{l}
x_d = x \, f(r) = x \, (1 + k_1 r + k_2 r^2)  \\
y_d = y \, f(r) = y \, (1 + k_1 r + k_2 r^2).
\end{array}
\end{eqnarray}
Again, let $c = y_d/x_d = y /x$. We have $y = cx$ where $c$ is a constant. Substituting $y = cx$ into the above equation gives
\begin{eqnarray}
\label{eqn: distortion model 3}
x_d &=& x \, \left[ 1+k_1 \sqrt{x^2 + c^2x^2} + k_2(x^2 + c^2x^2)\right] \nonumber\\
   &=& x \, \left[ 1+k_1 \sqrt{1 + c^2} \, {\tt sgn}(x) x + k_2(1 + c^2)x^2 \right] \nonumber\\
   &=& x + k_1 \sqrt{1 + c^2} \, {\tt sgn}(x) \, x^2 + k_2(1 + c^2) \, x^3,
\end{eqnarray}
where ${\tt sgn}(x)$ gives the sign of $x$. 
Let 
\begin{eqnarray}
f(x) = x + k_1 \sqrt{1 + c^2} \, {\tt sgn}(x) \, x^2 + k_2(1 + c^2) \, x^3. \nonumber
\end{eqnarray}
Clearly, $f(x)$ is also an odd function.

To perform the radial undistortion using the new distortion model in (\ref{eqn: new distortion model}), that is to extract $x$ from $x_d$ in (\ref{eqn: distortion model 3}), the following algorithm is applied:\\
\begin{itemize}

\item[{\rm \bf 1)}]{$x = 0$ iff $x_d = 0$,}

\item[{\rm \bf 2)}]Assuming that $x > 0$, (\ref{eqn: distortion model 3}) becomes
\begin{eqnarray}
x_d = x + k_1 \sqrt{1 + c^2} \, x^2 + k_2(1 + c^2) \, x^3. \nonumber
\end{eqnarray}
Using {\tt solve}, a Matlab Symbolic Toolbox function, we can get three possible solutions for the above equation 
denoted by $x_{1+}$, $x_{2+}$, and $x_{3+}$ respectively. To make the equations simple, let $y = x_d$, $p = k_1 \sqrt{1+c^2}$ and $q = k_2 (1+c^2)$. The three possible solutions for $y = x + p x^2 + qx^3$ are
{\small
\begin{eqnarray}
\label{eqn: three solutions}
x_{1+} &=& \frac{1}{6q}E_1 + \frac{2}{3} E_2 - \frac{p}{3q}, \nonumber\\
x_{2+} &=& -\frac{1}{12q} E_1 - \frac{1}{3} E_2 - \frac{p}{3q} + \frac{\sqrt{3}}{2}(\frac{1}{6q}E_1 - \frac{2}{3} E_2) \, {\bf \it j}, \\
x_{3+} &=& -\frac{1}{12q} E_1 - \frac{1}{3} E_2 - \frac{p}{3q} - \frac{\sqrt{3}}{2}(\frac{1}{6q}E_1 - \frac{2}{3} E_2) \, {\bf \it j}, \nonumber
\end{eqnarray}}
where
{\small
\begin{eqnarray}
\label{eqn: E1 E2}
E_1 &=& \{ 36pq+108yq^2-8p^3\nonumber\\
&& +12\sqrt{3}q\sqrt{4q-p^2+18pqy+27y^2q^2-4yp^3} \}^{1/3},\\
E_2 &=& \frac{p^2-3q}{qE_1}, \quad {\bf \it j} = \sqrt{-1}.  \nonumber
\end{eqnarray}}
From the above three possible solutions, we discard those whose imaginary parts are not equal to zero. Then, from the remaining, discard those solutions that conflict with the assumption that $x > 0$. Finally, we get the candidate solution $x_+$ by choosing the one closest to $x_d$ if the number of remaining solutions is greater than 1.

\item[{\rm \bf 3)}] Assuming that $x < 0$, there are also three possible solutions for 
\begin{eqnarray}
\label{eqn: minus case}
x_d = x - k_1 \sqrt{1 + c^2} \, x^2 + k_2(1 + c^2) \, x^3,
\end{eqnarray}
which can be written as 
\begin{eqnarray}
\label{eqn: minus case yqp}
y = x + (-p) x^2 + qx^3. 
\end{eqnarray}
The three solutions for (\ref{eqn: minus case yqp}) can thus be calculated from (\ref{eqn: three solutions}) and (\ref{eqn: E1 E2}) by substituting $p = -p$. With a similar procedure as described in the case for $x > 0$, we will have another candidate solution $x_-$. 

\item[{\rm \bf 4)}] Choose among $x_+$ and $x_-$ for the final solution of $x$ by taking the one closest to $x_d$.
\end{itemize}

The basic idea to extract $x$ from $x_d$ in (\ref{eqn: distortion model 3}) is to choose from several candidate solutions, whose analytical formula are known. The benefits of using this new radial distortion model are as follows:
\begin{itemize}
\item{Low order fitting, better for fixed-point implementation;}
\item{Explicit or analytical inverse function with no numerical iterations;}
\item{Better accuracy than using the radial distortion model $f(r) = 1 + k_1 r^2$.}
\end{itemize}

\thispagestyle{empty}
\section{Experimental Results and Comparisons}
\label{sec: Comparison Results}

\begin{figure}[htb]
\centering
\includegraphics[width=0.32\textwidth]{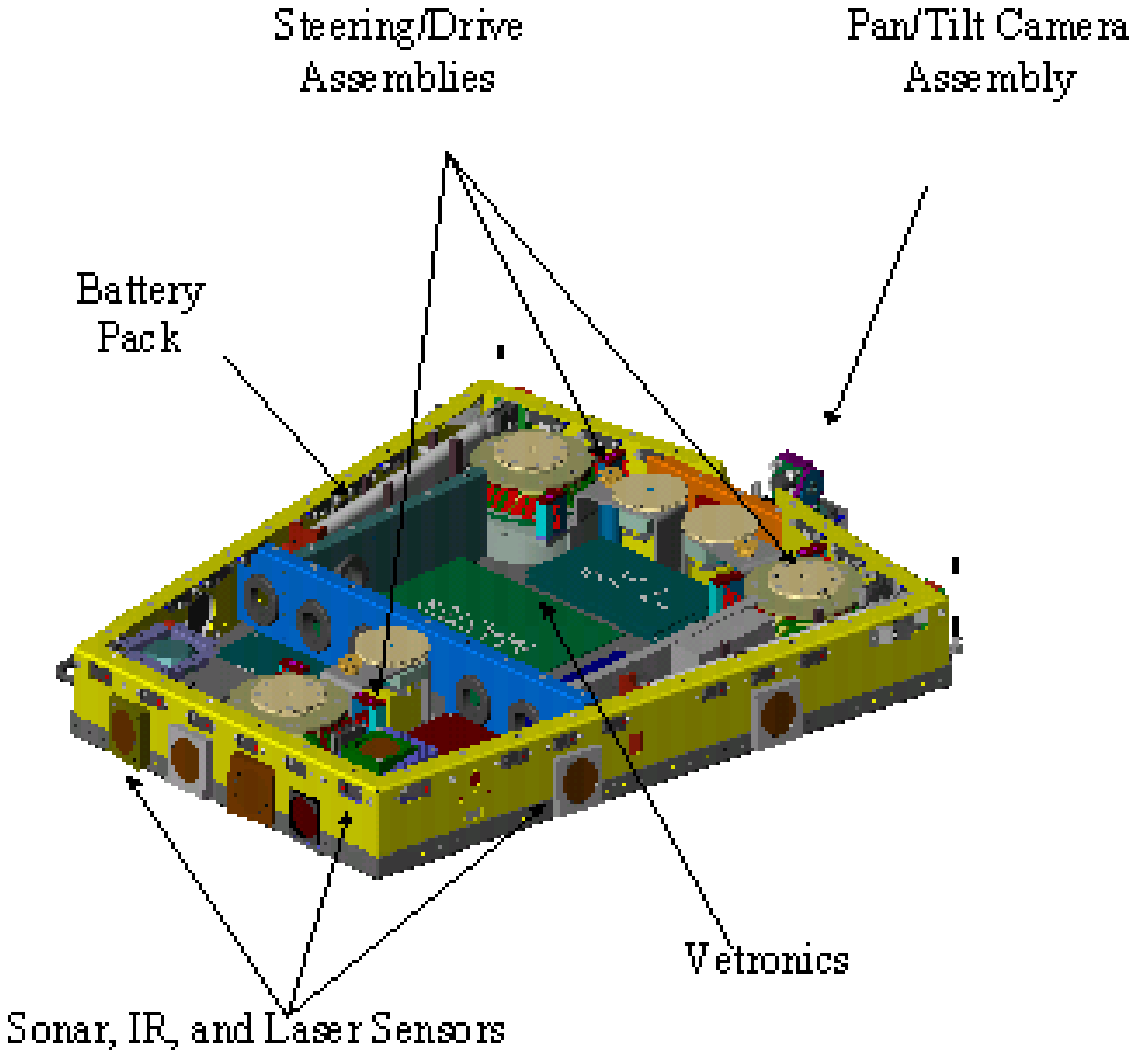}
\caption {The mechanical and vetronics layout of ODIS}
\label{fig: odis1}
\end{figure}

Now, we want to compare the performance of three different radial distortion models based on the final value of objective function after nonlinear optimization by the Matlab function {\tt fminunc}. The three different distortion models for comparison are:
\begin{eqnarray}
{\tt distortion \, model_1:} && f(r) = 1 + k_1 r^2 + k_2 r^4,      \nonumber\\
{\tt distortion \, model_2:} && f(r) = 1 + k_1 r^2,         		\nonumber\\
{\tt distortion \, model_3:} && f(r) = 1 + k_1 r + k_2 r^2.        \nonumber
\end{eqnarray}
Using the public domain test images \cite{zhang98calibrationwebpage}, the desktop camera images \cite{Lilicalreport02} (a color camera in our CSOIS), and the ODIS camera images \cite{Lilicalreport02} (the camera on ODIS robot built in our CSOIS, see Sec. \ref{sec: odis} and Fig. \ref{fig: odis1}), the final objective function ($J$), the 5 estimated intrinsic parameters ($\alpha, \beta, \gamma, u_0, v_0$), and the estimated distortion coefficients ($k_1, k_2$) are shown in Tables \ref{table: Comparison of Three Distortion Models1}, \ref{table: Comparison of Three Distortion Models2}, and \ref{table: Comparison of Three Distortion Models3} respectively \cite{Lilicalreport02}. The results show that the objective function of {\tt model$_3$} is always greater than that of {\tt model$_1$}, but much smaller than that of {\tt model$_2$}, which is consistent with our expectation. Note that, when doing nonlinear optimization with different distortion models, we always use the same exit thresholds.

To make the results in this paper repeatable by other researchers for further investigation, we present the options we use for the nonlinear optimization: \texttt{options = optimset(`Display', `iter', `LargeScale', `off', `MaxFunEvals', 8000, `TolX', $10^{-5}$,  `TolFun', $10^{-5}$, `MaxIter', 120)}. The raw data of the extracted feature locations in the image plane are also available upon request. 

A second look at the results reveals that for the camera used in \cite{zhang99calibrationinpaper,zhang99calibrationinreport,zhang98calibrationwebpage}, which has a small lens distortion, the advantage of {\tt model$_3$} over {\tt model$_2$} is not so significant. When the cameras are experiencing severe distortion, the radial distortion {\tt model$_3$} gives a much better performance over {\tt model$_2$}, as can be seen from Tables \ref{table: Comparison of Three Distortion Models2} and \ref{table: Comparison of Three Distortion Models3}.

{\small
\begin{table}[htb]
\begin{minipage}{3.5in}
\setcounter{mpfootnote}{\value{footnote}}
\renewcommand{\thempfootnote}{\fnsymbol{mpfootnote}}
\centering
\caption[Comparison of Distortion Models Using Microsoft Images]%
{Comparison of Distortion Models Using Images in \cite{zhang98calibrationwebpage} 
\footnote{($k_1, k_2$) for {\tt model$_1$} and {\tt model$_2$} are defined in (\ref{eqn: general distortion model}) and ($k_1, k_2$) for {\tt model$_3$} is defined in (\ref{eqn: new distortion model}).}}
\label{table: Comparison of Three Distortion Models1}
\renewcommand{\arraystretch}{1}
    {\begin {tabular}{|c|r|r|r|}\hline
    \multicolumn{4}{|c|}{\bf Microsoft Images}\\\hline
    {\bf Model} & $\#1$  & $\#2$ & $\#3$ \\\hline
    {\bf $J$} &  {\bf 144.88}   & {\bf 148.279}  & {\bf 145.659} \\\hline
    $\alpha$ &  832.5010 & 830.7340 & 833.6623  \\
    $\gamma$ &    0.2046 &   0.2167 & 0.2074    \\
    $u_0$    &  303.9584 & 303.9583 & 303.9771  \\
    $\beta$  &  832.5309 & 830.7898 & 833.6982  \\
    $v_0$    &  206.5879 & 206.5692 & 206.5520  \\
    $k_1$    &   -0.2286 &  -0.1984 &  -0.0215  \\
    $k_2$    &    0.1903 &        0 &  -0.1565  \\\hline
    \end {tabular}}
\end{minipage}
\end{table}}

{\small
\begin{table}[htb]
\centering
\caption[Comparison of Distortion Models Using Desktop Images]%
{\small Comparison of Distortion Models Using Desktop Images in \cite{Lilicalreport02}}
\label{table: Comparison of Three Distortion Models2}
\renewcommand{\arraystretch}{1}
{\begin {tabular}{|c|r|r|r|}\hline
\multicolumn{4}{|c|}{\bf Desktop Images}\\\hline
{\bf Model} & $\# 1$    & $\# 2$ & $\# 3$ \\\hline
{\bf $J$}&  {\bf 778.9768}   &  {\bf 904.68}  & {\bf 803.307}\\\hline
$\alpha$ & 277.1457   & 275.5959 & 282.5664  \\
$\gamma$ &    -0.5730 &  -0.6665 &   -0.6201 \\
$u_0$    &   153.9923 & 158.2014 &  154.4891 \\
$\beta$  &   270.5592 & 269.2307 &  275.9040 \\
$v_0$    &   119.8090 & 121.5254 &  120.0952 \\
$k_1$    &    -0.3435 &  -0.2765 &   -0.1067 \\
$k_2$    &     0.1232 &        0 &   -0.1577 \\\hline
\end {tabular}}
\end{table}}

{\small
\begin{table}[htb]
\centering
\caption[Comparison of Distortion Models Using ODIS Images]%
{Comparison of Distortion Models Using ODIS Images in \cite{Lilicalreport02}}
\label{table: Comparison of Three Distortion Models3}
\renewcommand{\arraystretch}{1}
{\begin {tabular}{|c|r|r|r|}\hline
\multicolumn{4}{|c|}{\bf ODIS Images} \\\hline
{\bf Model} & $\# 1$ & $\# 2$ & $\# 3$ \\\hline
{\bf $J$} & {\bf 840.2650}  &  {\bf 933.098} & {\bf 851.262}  \\\hline
$\alpha$ &260.7636   & 258.3206    & 266.0861 \\
$\gamma$ &   -0.2739 &    -0.5166  &  -0.3677 \\
$u_0$    &  140.0564 &   137.2155  & 139.9177 \\
$\beta$  &  255.1465 &   252.6869  & 260.3145 \\
$v_0$    &  113.1723 &   115.9295  & 113.2417 \\
$k_1$    &   -0.3554 &    -0.2752  &  -0.1192 \\
$k_2$    &    0.1633 &          0  &  -0.1365 \\\hline
\end {tabular}}
\end{table}}

\thispagestyle{empty}
\section{Application: Non-iterative Yellow Line Alignment with a Calibrated Camera on ODIS}
\label{sec: Application}

\subsection{What is ODIS?}
\label{sec: odis}
The Utah State University Omni-Directional Inspection System) (USU ODIS) is a small, man-portable mobile robotic system that can be used for autonomous or semi-autonomous inspection under vehicles in a parking area \cite{mooreTA01,mooreFSR01,lili02visualservo}. The robot features (a) three ``smart wheels'' \cite{moore_csm} in which both the speed and direction of the wheel can be independently controlled through dedicated processors, (b) a vehicle electronic capability that includes multiple processors, and (c) a sensor array with a laser, sonar and IR sensors, and a video camera. A unique feature in ODIS is the notion of the ``smart wheel'' developed by the Center for Self-Organizing and Intelligent Systems (CSOIS) at USU which has resulted in the so-called T-series of omni-directional (ODV) robots \cite{moore_csm}. With the ODV technique, our robots including ODIS, can achieve complete control of the vehicle's orientation and motion in a plane, thus making the robots almost holonomic - hence ``omni-directional''. ODIS employs a novel parameterized command language for intelligent behavior generation \cite{mooreFSR01}. A key feature of the ODIS control system is the use of an object recognition system that fits models to sensor data. These models are then used as input parameters to the motion and behavior control commands \cite{mooreTA01}. Fig.~\ref{fig: odis1} shows the mechanical layout of the ODIS robot. The robot is 9.8 cm tall and weighs approximately 20 kgs. 

\subsection{Motivation}
The motivation to do camera calibration and radial undistortion is to better serve the wireless visual servoing task for ODIS. Our goal is to align the robot to a parking lot yellow line for localization. Instead of our previous yellow line alignment methods described in \cite{lili02visualservo,berkemeier01visualservo}, we can align to the yellow line with a non-iterative way using a calibrated camera. The detail procedure is discussed in the next section.

\subsection{Localization Procedure}
Let us begin with a case when only ODIS's yaw and $x, y$ positions are unknown while ODIS camera's pan/tilt angles are unchanged since calibration. The task of yellow line alignment is described in detail as follows:
\begin{itemize}
\item {\bf Given:}
\begin{itemize}
\item{3D locations of yellow line's two ending points}
\item{Observed ending points of yellow line in the image plane using ODIS camera}
\item{ODIS camera's pan/tilt angles}
\item{ODIS camera's intrinsic parameters}
\item{Radial distortion model and coefficients}
\end{itemize}
\item{\bf Find:}
ODIS's actual yaw and $x, y$ positions
\end{itemize}

\begin{figure}[htb]
\centering
\includegraphics[width=0.45\textwidth]{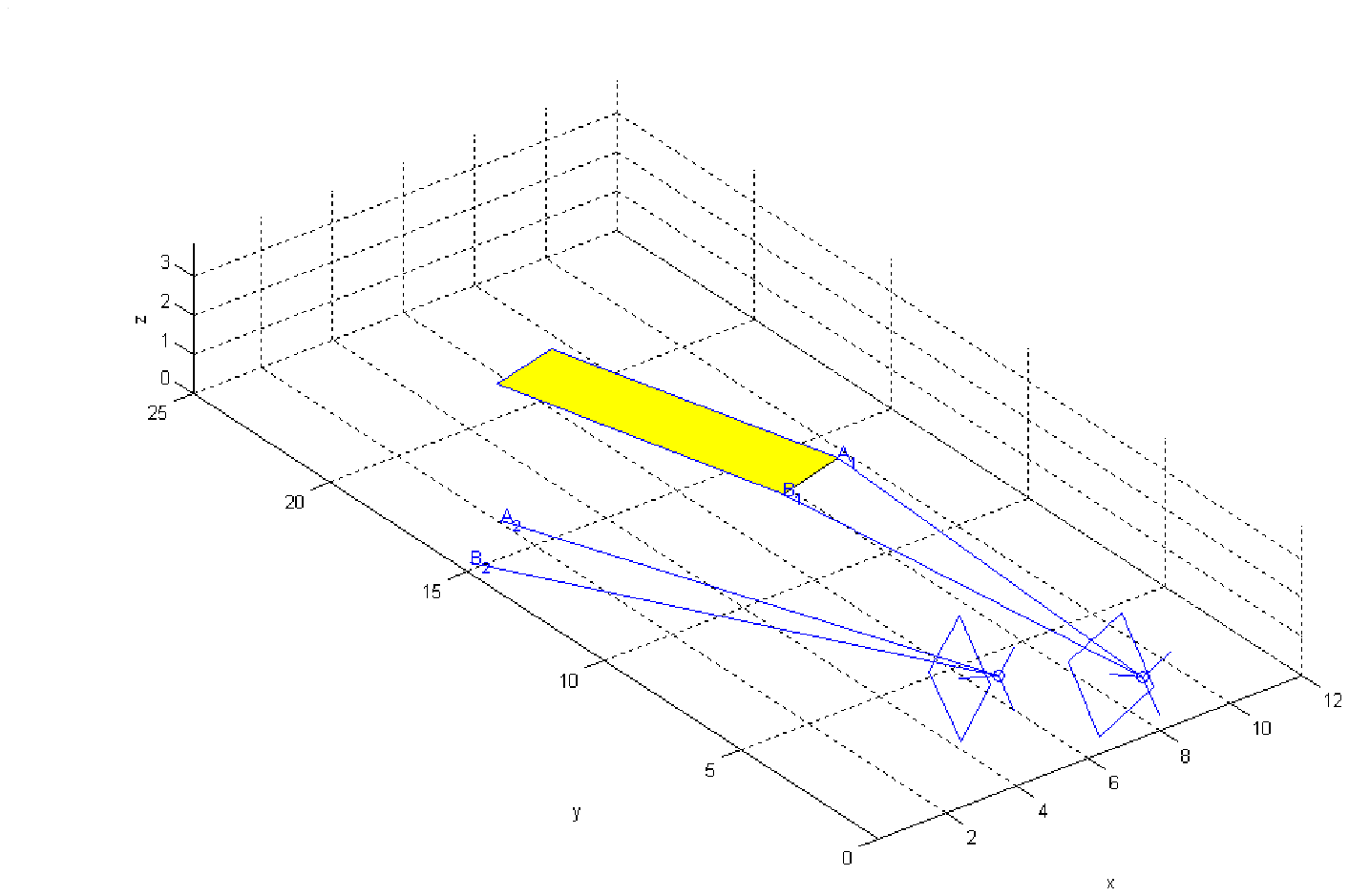}
\caption {The task of yellow line alignment}
\label{fig: visual servo1}
\end{figure}

Knowing that a change in ODIS's yaw angle only results in a change of angle $s$ in the $ZYZ$ Euler angles $(a,b,s)$. So, when using $ZYZ$ Euler angles to identify ODIS camera's orientation, the first two variables $a, b$ are unchanged. In Fig. \ref{fig: visual servo1}, after some time of navigation, the robot thinks it is at Position 2, but actually at Position 1. Then it sees the yellow line, whose locations in 3D world reference frame are known from map (denoted by ${P_A}_1^w$ and ${P_B}_1^w$). After extracting the corresponding points in the image plane of the yellow line's two ending points, we can calculate the undistorted image points and thus recover the 3D locations of the two ending points (denoted by ${P_A}_2^w$ and ${P_B}_2^w$), using ODIS camera's 5 intrinsic parameters and radial distortion coefficients. From the difference between the yellow line's actual locations in map and the recovered locations, the deviation in the robot's $x, y$ positions and yaw angle can be calculated.

Let $(x, y)$ be the undistorted points in the camera frame corresponding to the yellow line's two ending points in the 3D world frame. Let $R_2$ and ${\bf t}_2$ be the rotation matrix and translation vector at position 2 (where the vehicle thinks it is at), similarly $R_1$ and ${\bf t_1}$ at position 1 (the true position and orientation), we can write $R_2 = \Delta R \cdot R_1$ and ${\bf t}_2 = {\bf t}_1 + \Delta {\bf t}$, where $\Delta R$ and $\Delta {\bf t}$ are the deviation in orientation and translation. If the transform from the world reference frame to the camera frame is $P^c = R^{-1} (P^w - {\bf t})$, first we can calculate ${P_A}_2^w$ and ${P_B}_2^w$.

Let ${P_A}_2^w = [{X_A}_2^w, {Y_A}_2^w, 0]$, we have
\begin{eqnarray}
\left [\matrix{X^c \cr Y^c \cr Z^c }\right] = R_2^{-1} \left[\matrix{{X_A}_2^w - {\bf t}_{21}\cr {Y_A}_2^w - {\bf t}_{22}\cr -{\bf t}_{23}}\right].
\end{eqnarray}
Since \begin{eqnarray}
\frac{X^c}{x} = \frac{Y^c}{y} = \frac{Z^c}{1},
\end{eqnarray}
we have two equations containing two variables and ${P_A}_2^w$ can be calculated out. By the same way, we can get ${P_B}_2^w$. \\
Once ${P_A}_2^w$ and ${P_B}_2^w$ are known, we have
\begin{eqnarray}
\label{eqn: point A}
\lambda \left [ \matrix {
x \cr
y \cr
1
} \right] = R_2^{-1} \Delta R ({P_A}_1^w - {\bf t}_1) = R_2^{-1} ({P_A}_2^w - {\bf t}_2),
\end{eqnarray}
where $\lambda$ is a scaling factor.
From (\ref{eqn: point A}), we get
\begin{eqnarray}
\label{eqn: A - AA}
R_2^{-1} [\Delta R ({P_A}_1^w - {\bf t}_1) - {P_A}_2^w + {\bf t}_2] = 0.
\end{eqnarray}
Similarly, we get
\begin{eqnarray}
\label{eqn: B - BB}
R_2^{-1} [\Delta R ({P_B}_1^w - {\bf t}_1) - {P_B}_2^w + {\bf t}_2] = 0.
\end{eqnarray}
Using the above two equations, we have $({P_A}_2^w - {P_B}_2^w) = \Delta R \, ({P_A}_1^w - {P_B}_1^w)$, where $\Delta R$ is of the form
\begin{eqnarray}
\Delta R = \left [ \matrix {
\cos(\Delta \theta) & -\sin(\Delta \theta) & 0 \cr
\sin(\Delta \theta) &  \cos(\Delta \theta) & 0 \cr
0 & 0 & 1
} \right].
\end{eqnarray}
So, $\Delta \theta$ is just the rotation angle from vector ${P_A}_1^w \rightarrow {P_B}_1^w$ to vector ${P_A}_2^w \rightarrow {P_B}_2^w$. When $\Delta R$ is available, ${\bf t}_1$ can be calculated as ${\bf t}_1 = {P_A}_1^w - {\Delta R}^{-1}({P_A}_2^w -{\bf t}_2)$.

\thispagestyle{empty}
\section{Concluding Remarks}
\label{sec: Conclusion}


This paper proposes a new radial distortion model that belongs to the polynomial approximation category. The appealing part of this distortion model is that it preserves high accuracy together with an easy analytical undistortion formula. Experiments results are presented showing that this distortion model is quite accurate and efficient especially when the actual distortion is significant. An application of the new radial distortion model is non-iterative yellow line alignment with a calibrated camera on ODIS.

\bibliography{calibration,csois1,csois2}
\end{document}